\documentclass[runningheads]{llncs}

\usepackage{textcomp}
\usepackage{amsmath}
\usepackage[T1]{fontenc}
\usepackage{multirow}
\usepackage{caption}
\usepackage{float} 
\usepackage{array}          
\usepackage{tabularx}       
\usepackage{tabu}       
\usepackage{tabularray}
\usepackage{wasysym}
\usepackage{pifont}
\usepackage{verbatim}

\usepackage{graphicx}
\begin{document}
%
\title{MuCoS: Efficient Drug-Target Prediction through Multi-Context-Aware Sampling}
%
%
\author{Haji Gul\inst{1a}\orcidID{0000-0002-2227-6564} \and
Ajaz Ahmad Bhat\inst{1b}\orcidID{0000-0002-6992-8224} \and
Abdul Ghani Haji Naim\inst{1c}\orcidID{0000--0000-0000-0000}}

\authorrunning{H. Gul et al.}
%
\institute{$^1$Universiti Brunei Darussalam, Jalan Tungku Link, Gadong BE1410,
\email{($^a$23h1710@ubd, $^b$ajaz.bhat@ubd, $^c$ghani.naim@ubd).edu.bn}\\
}
\maketitle              
\begin{abstract}
Drug-target interactions are critical for understanding biological processes and advancing drug discovery. However, traditional methods such as ComplEx-SE, TransE, and DistMult struggle with unseen relationships and negative triplets, which limits their effectiveness in drug-target prediction. To address these challenges, we propose Multi-Context-Aware Sampling (MuCoS), an efficient and positively accurate method for drug-target prediction. MuCoS reduces computational complexity by prioritizing neighbors of higher density to capture informative structural patterns. These optimized neighborhood representations are integrated with BERT, enabling contextualized embeddings for accurate prediction of missing relationships or tail entities. MuCoS avoids the need for negative triplet sampling, reducing computation while improving performance over unseen entities and relations. Experiments on the KEGG50k biomedical dataset show that MuCoS improved over existing models by 13\% on MRR, 7\% on Hits@1, 4\% on Hits@3, and 18\% on Hits@10 for the general relationship, and by 6\% on MRR, 1\% on Hits@1, 3\% on Hits@3, and 12\% on Hits@10 for prediction of drug-target relationship.
\end{abstract}
%
\section{Introduction}
Complex biological entities such as drugs, proteins, diseases, and genes, along with their interactions, form a critical foundation for understanding biological systems. In this context, the prediction of drug-target interactions (DTIs) is a key task that aims to identify missing interactions between drugs and their target proteins. Accurate prediction of DTI accelerates the drug discovery process and reduces associated costs \cite{sachdev2019comprehensive}. It also enables researchers to prioritize potential drugs and gain insight into the mechanisms of drug activity. Resources such as KEGG50k \cite{caivio2021diversity} and Hetionet \cite{ramosaj2023reasoning} provide structured representations of biological components, including pathways, genes, drugs, and diseases, along with their associations. For example, KEGG50k illustrates relationships such as pathway genes, drug target genes, and disease genes, which are essential for advancing biomedical research and drug repurposing.

A study uses a computation-based approach to solve a drug interaction prediction challenge in a medical knowledge graph (KG) that involves drugs, proteins, diseases, and pathways \cite{mohamed2019drug}. The model, ComplEx-SE, uses a modified version of the advanced ComplEx KGE model, which uses a squared error-based loss function for more accurate predictions. The model is trained and assessed using a medical KG dataset, KEGG50k, sourced from the KEGG database. The study emphasizes the general prediction of drug targets, which involves discovering possible relationships between drugs and their target proteins. However, the model faces challenges with unseen relationships and entities absent from the training data, limiting its ability to predict interactions involving novel drugs or targets. In addition, due to the cost and time demands of traditional pharmacological assessments for drug-target interaction (DTI) prediction, there is a significant demand for accurate computational methods to determine DTIs  \cite{agu2024piquing}. 

We propose a novel MuCoS method that overcomes these limitations by collecting contextual information from adjacent entities and their relationships. The context is subsequently aggregated and used in the BERT model to enhance the prediction of the relationship and the entities. This research advances the medical domain in the following ways:
\begin{itemize}
    \item \textbf{Contextualized Relation Prediction}: We propose MuCoS, which uses head-and-tail-optimized neighboring contextual structural features to predict general and drug-target relationships and improve predictive performance.
    \item\textbf{Contextualized Tail Prediction}: Predicted tail entities using contextual head and relationship-optimized neighboring structural information for tail prediction in general and drug-target datasets.
    \item \textbf{Description-Independence}: Unlike previous approaches, MuCoS does not require entity descriptions.
    \item \textbf{Negative Sample-Free Training}: Using cross-entropy loss, MuCoS-KGC eliminates the need for negative sampling, improving training efficiency and robustness.
\end{itemize}

\section{Related Work}
Similarity-based DTI prediction methods utilize drug and target similarity measures in conjunction with the distance between each pair of drugs and their respective targets \cite{shi2018drug}. The similarity is obtained using a distance function like the Euclidean.
The DTI prediction using feature-based approaches uses mainly the support vector machine \cite{zhang2017drugrpe}. A pair of targets and drugs can be distinguished by features, leading to binary classification or two-class clustering based on positive or negative interactions.
Currently, graph-based approaches used in parallel with network-based methods to predict missing DTI are considered reliable interaction prediction methods \cite{ban2019nrlmfbeta}. Here, drug-drug similarity, target-target similarity, and known interactions between DTI are integrated into a heterogeneous network, operating on the simple logical principle that similar drugs interact with similar targets.

Embedding-based approaches (KGE) such as TransE \cite{bordes2013translating} and RotatE \cite{sun2019rotate} incorporate the fundamental semantics of a graph by expressing entities and relationships between them as continuous vectors. In contrast, semantic or textual matching techniques such as DistMult \cite{yang2014embedding}, utilize scoring functions to evaluate the similarity of entities and relationships. By looking at the scores for all the relationships and entities linked to a certain head/tail and relationship, these methods find the most likely relationship or tail entity during testing. 

In summary, existing KGC methodologies exhibit notable limitations. Embedding models such as TransE, RotatE, and others fail to generalize to unseen entities and relations because of their dependence on predefined embeddings, limiting adaptability to new structures. Text-based and LLM-based approaches depend on descriptive entity information, which is often sparse or inconsistent across datasets. In addition, embedding methods, as well as some text-based methods, rely on extensive negative sampling during training, leading to high computational costs, especially on large datasets. 
\section{Methodology}
This section presents the MuCoS technique, which predicts relations given a head-and-tail entity from a KEGG50K KG. MuCoS also predicts tail entities based on the head and relationship. We briefly describe the problem before illustrating the model and then explain in detail how to extract various (head, relation, and tail) components optimized for contextual information. After optimizing contextual information extraction, we explain how to pass them to the BERT model for relationship and tail prediction.  
KG is a structured representation of knowledge mathematically represented as $G = (\mathcal{E}, \mathcal{R}, \mathcal{T})$, where $\mathcal{E}$ denotes the set of entities, $\mathcal{R}$ represents the set of relations and $\mathcal{T} \subseteq \mathcal{E} \times \mathcal{R} \times \mathcal{E}$ is the set of triples. Each triple $(e_h, r, e_t)$ indicates a directed relationship $r$ from the head entity $e_h$ to the tail entity $e_t$. This work incorporated the prediction of two components of the KG relationship prediction $(e_h, ?, t)$, also known as link prediction and tail prediction $(e_h, r, ?)$, inferring the missing tail. 

Previously, we proposed the CAB-KGC model, which extracts full contextual information from both the head entity and the relationship \cite{gul2024contextualized}. It takes the head context (Hc) and relationship context (Rc) from the head entity and relationship and sets them up as input for a BERT classifier. However, the main constraint of CAB-KGC is that it is computationally expensive. An effective multi-context-aware sampling (MuCoS) method is presented in this work, which is faster than CAB-KGC. In some cases, the accuracy of MuCoS is lower, but we proved it is $\approx 10$ time faster than CAB-KGC.

\begin{figure}[htbp]
\includegraphics[width=\textwidth]{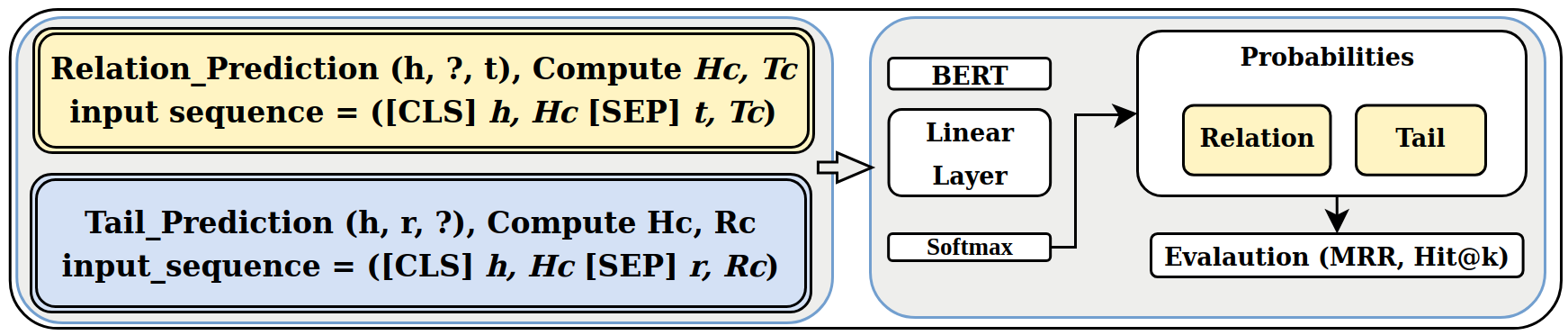}
\caption{\scriptsize A concise view of the MuCoS model pipeline for predicting the relation (given a head $h$ and a tail $t$ entity) and tail entity (given a head entity $h$ and a relationship $r$). The box on the left illustrates the input sequence to the BERT model, containing a union of $\mathcal{R}(h)$ (set of all relations) and $\mathcal{E}(h)$ (set of all neighboring entities). The BERT model, combined with a linear classifier and softmax, generates probabilities for tail entities.}
\label{fig:methodology}
\end{figure}

The MuCoS addresses the significance of contextual information derived from the head entity and relationship, integrating it with BERT. Given a query in the form \((h, ?, t)\), the framework extracts and optimizes contextual information from the head and tail entities, aggregates and processes it through BERT for the prediction of the relationship entity. Similarly, the model extracts head and relationship-optimized contextual information and predicts the tail entity $(h,r,?)$. A concise graphical view of MuCoS is reported in Figure \ref{fig:methodology}. The subsequent sections provide a detailed explanation of how to compute the context information for the head, tail, and relation, including mathematical representations and clear graphical illustrations.
\vspace{-15pt}

\begin{figure}[H]
    \centering
    \includegraphics[width=\textwidth]{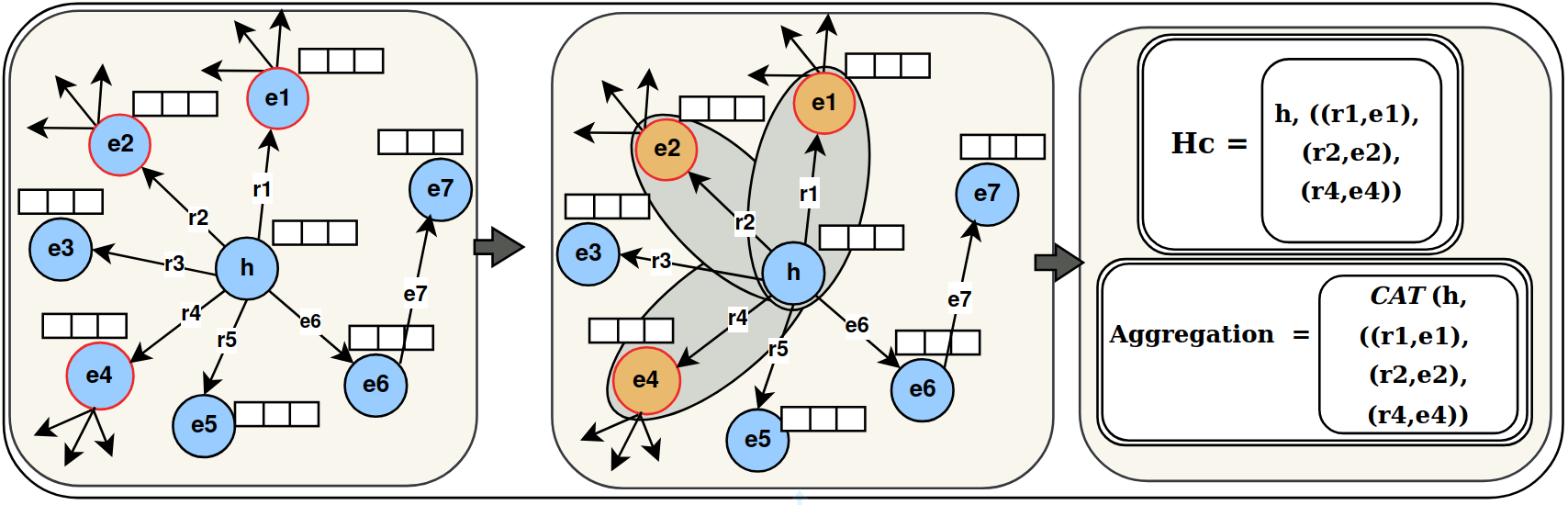}
    \caption{\scriptsize MuCoS \( \mathcal{H}_c \) construction. The left box illustrates the extraction of \( \mathcal{H}_c \), which consists of the set of relationships \( \mathcal{R}(h) \) (\( r_1, r_2, r_3, r_4, r_5, r_6 \)) and the set of neighboring entities \( \mathcal{E}(h) \) (\( e_1, e_2, e_3, e_4, e_5, e_6 \)) associated with the head entity \( h \). The middle box shows optimization, where each entity $e$ density $d(e)$ is computed based on its frequency in triples. The optimized head context $ H_c*$ includes the top-$k$ most influential entities and relationships. Lastly, the optimized \( \mathcal{H}_c^* \) is then aggregated into a unified representation \( \mathcal{H}_{\text{agg}} \) using concatenation, as shown in the right box.}
    \label{fig:hc}
\end{figure}

\begin{figure}[htbp]
    \centering
    \includegraphics[width=\textwidth]{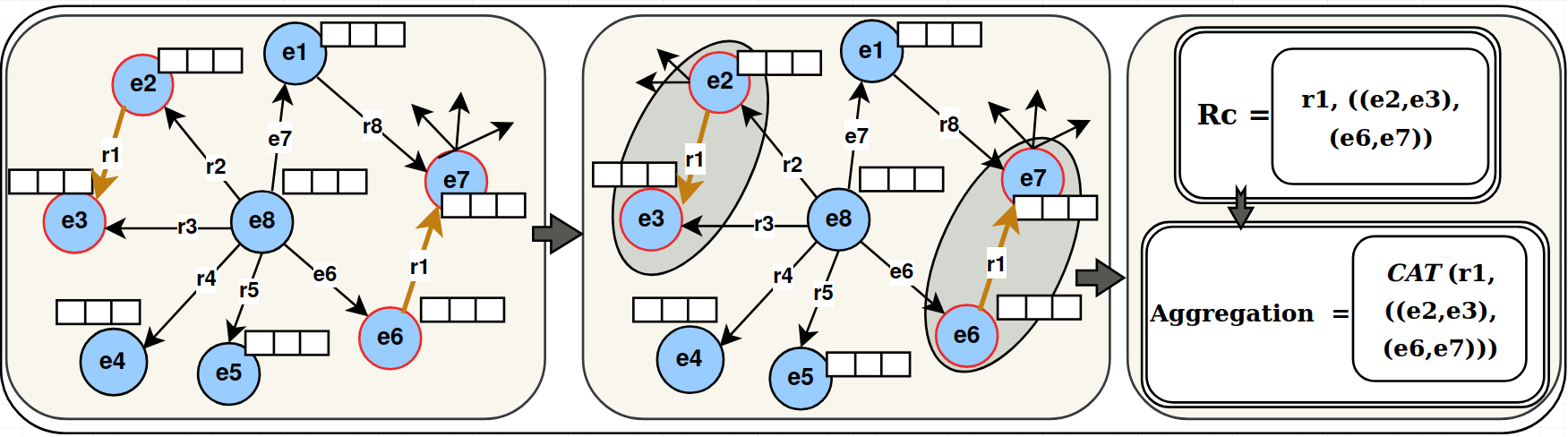}
    \caption{\scriptsize \( \mathcal{R}_c \) construction, the left section shows the appearance of $r_1$ and lists the entities \( \mathcal{E}(r)\) that are connected to the relationship \(r_1 \). The middle section is used for optimization; we suppose that the current top $k$ for $r$ selection is two, so retaining both relevant entity pairs \((e_2, e_3)\) and \((e_6, e_7)\). The optimized relationship context \( \mathcal{R}_c^* \) is aggregated \( \mathcal{R}_{\text{agg}} \) using concatenation, as depicted in the right box. }
    \label{fig:rc}
\end{figure}

\begin{figure}[H]
    \centering
    \includegraphics[width=\textwidth]{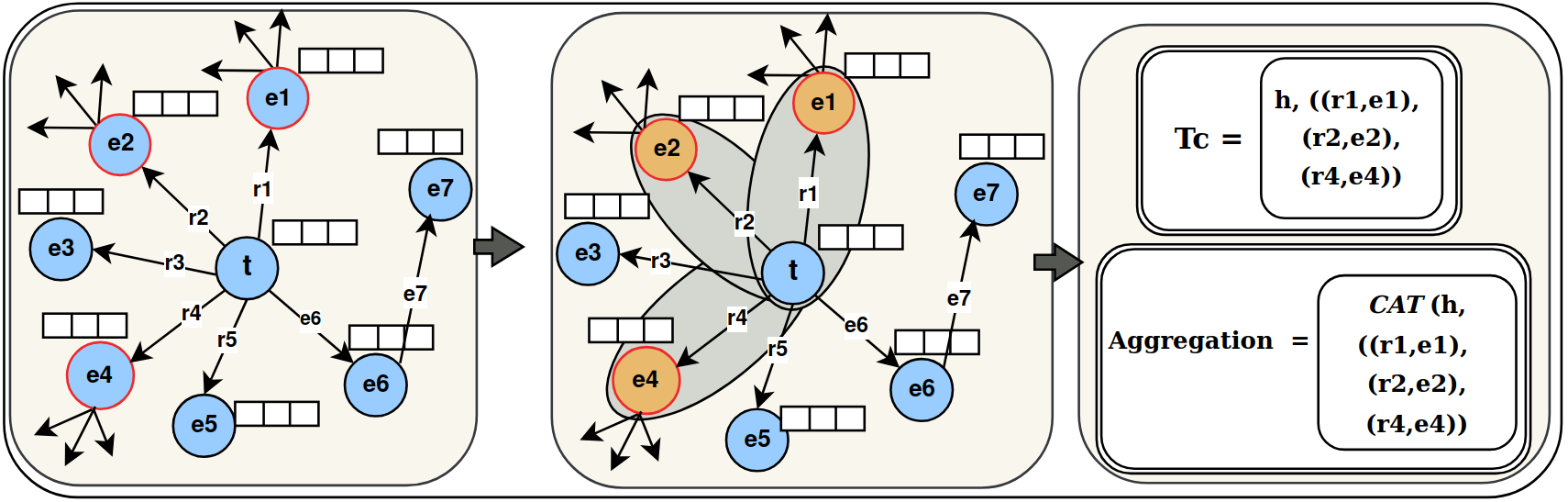}
    \caption{\scriptsize Similarly, the \textit{left box} represents the extraction of \( \mathcal{T}_c \) the set of relationships \( \mathcal{R}(t) \) (\( r_1, r_2, r_3, r_4, r_5, r_6 \)) and neighboring entities \( \mathcal{E}(t) \) (\( e_1, e_2, e_3, e_4, e_5, e_6 \)) associated with the tail entity \( t \). The middle box illustrates the optimization step, where the density $d(e)$ of each entity $e$ is calculated based on its frequency in triples. The top-$k$ most influential entities and relationships are selected to form the optimized tail context $T_c^*$. The concatenation results in a unified representation $T_{\text{agg}}$. }
    \label{fig:tc}
\end{figure}

\vspace{-15pt}
To derive the \textit{\textbf{head context}} \(\mathcal{H}_c\), we first identify the relationships \(\mathcal{R}(h)\) associated with the head entity \( h \). Next, we extract the entities \(\mathcal{E}(h)\) directly connected \( h \) through these relationships. Mathematically, this is represented as follows:
\begin{equation}
\mathcal{R}(h) = \{ r_1, r_2, \ldots, r_k \}, ~ ~ \mathcal{E}(h) = \{ e_1, e_2, \ldots, e_m \}
\label{eq:rh}
\end{equation}
The head context \(\mathcal{H}_c\) is constructed by taking the union of the relationships and entities associated with the head entity:
\begin{equation}
\mathcal{H}_c = \mathcal{R}(h) \cup \mathcal{E}(h)
\label{eq:hc}
\end{equation}
To refine the head context, we introduce a density-based optimization step. The density \( d(e) \) of an entity \( e \) is defined as the frequency of its appearance in triples in KEGG50K. For each entity \( e \in \mathcal{E}(h) \), the density is computed as:
\begin{equation}
d(e) = |\{ (e, r, t) \in \mathcal{G} \text{ or } (h, r, e) \in \mathcal{G} \}|
\label{eq:density}
\end{equation}
Using the density metric, we select the top-\( k \) most influential entities and relationships:
\begin{equation}
\text{top}_k(\mathcal{E}(h)) = \text{sort}(\mathcal{E}(h), \text{by } d(e))[:k]
\label{eq:topk_eh}
\end{equation}
\begin{equation}
\text{top}_k(\mathcal{R}(h)) = \text{sort}(\mathcal{R}(h), \text{by } d(e))[:k]
\label{eq:topk_rh}
\end{equation}
The optimized head context \(\mathcal{H}_c^*\) is then defined and subsequently concatenated to form a unified representation of the head entity and its neighborhood. For more  clarification and detail, see Figure \ref{fig:hc}. 
\begin{equation}
\mathcal{H}_c^* = \text{top}_k(\mathcal{R}(h)) \cup \text{top}_k(\mathcal{E}(h)), \Longrightarrow \mathcal{H}_{\text{agg}} = \text{CAT}(\mathcal{H}_c^*)
\label{eq:ohc}
\end{equation}
\begin{equation}
\mathcal{H}_{\text{agg}} = \text{CAT}(\mathcal{H}_c^*)
\label{eq:agg_h}
\end{equation} 
\vspace{-15pt}
The\textit{\textbf{ relationship context}} \(\mathcal{R}_c\) is derived by identifying the entities \(\mathcal{E}(r)\) associated with the relationship \( r \):
\begin{equation}
\mathcal{E}(r) = \{ e_1, e_2, \ldots, e_l \}, ~ \mathcal{R}_c = \mathcal{E}(r)
\label{eq:er}
\end{equation}
Similar to the head context, the relationship context is optimized by selecting the top-\( k 
\) entities based on their density:
\begin{equation}
\text{top}_k(\mathcal{E}(r)) = \text{sort}(\mathcal{E}(r), \text{by } d(e))[:k]
\label{eq:topk_er}
\end{equation}
The optimized relationship context \(\mathcal{R}_c^*\) is aggregated through concatenation to represent the relationship and its associated entities, given in Equation \ref{eq:orc}. Figure \ref{fig:rc} provides a detailed graphical view of the relation context. 
\begin{equation}
\mathcal{R}_c^* = \text{top}_k(\mathcal{E}(r)), ~ \Longrightarrow \mathcal{R}_{\text{agg}} = \text{CAT}(\mathcal{R}_c^*)
\label{eq:orc}
\end{equation}
Similarly, to extract the \textit{\textbf{tail context}} \(\mathcal{T}_c\), the following formulation is presented, which extracts and optimizes contextual information associated with the tail entity. The graphical representation is given in Figure \ref{fig:tc}.
\begin{equation}
\mathcal{R}(t) = { r_1, r_2, \ldots, r_k }, \quad \mathcal{E}(t) = { e_1, e_2, \ldots, e_m }
\end{equation}
\begin{equation}
\mathcal{T}_c = \mathcal{R}(t) \cup \mathcal{E}(t)
\end{equation}
\begin{equation}
d(e) = |{ (h, r, e) \in \mathcal{G} \text{ or } (e, r, t) \in \mathcal{G} }|
\end{equation}
\begin{equation}
\text{top}_k(\mathcal{E}(t)) = \text{sort}(\mathcal{E}(t), \text{by } d(e))[:k]  \text{by } d(e))[:k]
\end{equation}
\begin{equation}
\text{top}_k(\mathcal{R}(t)) = \text{sort}(\mathcal{R}(t), \text{by } d(e))[:k]
\end{equation}
\begin{equation}
\mathcal{T}_c^* = \text{top}_k(\mathcal{R}(t)) \cup \text{top}_k(\mathcal{E}(t))
\end{equation}
\begin{equation}
\mathcal{T}_{\text{agg}} = \text{CAT}(\mathcal{T}_c^*)
\end{equation}
\begin{itemize}
    \item For \textbf{relation prediction} \((h, ?, t)\), the concatenated representations \(\mathcal{H}_{\text{agg}}\) (head context) and 
    \(\mathcal{T}_{\text{agg}}\) (tail context) are combined with the head entity \( h \) and tail entity \( t \) to form the input sequence for BERT. The input sequence \([h, \mathcal{H}_{\text{agg}}, t, \mathcal{T}_{\text{agg}}]\) passes through its transformer layers, generating a contextualized representation for each token. A classification layer is added on top of BERT to predict the relationship \( r \). The probability distribution over all possible relationships is computed using the softmax function:
\begin{equation}
    P(r \mid h, t) = \text{softmax}(W \cdot \text{BERT}(h, \mathcal{H}_{\text{agg}}, t, \mathcal{T}_{\text{agg}}))
\end{equation}

The model is trained using cross-entropy loss, which compares the predicted probability distribution with the true label for the relationship. The loss function is defined as
\begin{equation}
    L = - \sum_{i=1}^{N} y_i \log P(r_i \mid h, t),
\end{equation}
    \item For \textbf{tail prediction} \((h, r, ?)\), concatenated representations \(\mathcal{H}_{\text{agg}}\) and \(\mathcal{R}_{\text{agg}}\) are then combined with the head entity \(h\) and relationship \(r\). Mathematically, it  can be represented as:
 \begin{equation}
    P(t \mid h, r) = \text{softmax}(W \cdot \text{BERT}(h, \mathcal{H}_{\text{agg}}, r, \mathcal{R}_{\text{agg}}))
\end{equation}
The model is trained using cross-entropy loss, which compares the predicted probability distribution with the true label for the tail entity:
\begin{equation}
\mathcal{L} = -\sum_{i=1}^{N} y_i \log P(t_i \mid h, r)
\label{eq:loss}
\end{equation}
where \( y_i \) is the one-hot encoded true label for the tail entity \( t_i \), and \( P(t_i \mid h, r) \) is the predicted probability for \( t_i \).
\end{itemize}
\subsection{Experimental Setup}
\label{ex:det}
\textit{Dataset:} The proposed model was evaluated on the KEGG50k medical domain dataset, which consists of 63,080 triples overall, split into 57,080 triples for training, 3,000 for validation, and 3,000 for testing. These triples show relationships among multiple biological entities: pathways, genes, drugs,  and diseases. There are 16,201 unique nodes and 9 different types of relationships.

\textit{Hyperparameters:} The input sequence is tokenized using a maximum length of 128 tokens. Using an AdamW optimizer with a learning rate of 5e-5, the 50-epoch training process runs with a batch size of 16. The experiments were performed on an NVIDIA GeForce RTX 3090 GPU with 24 GB of memory.

\textit{Evaluation:} Various standard evaluations given in equation \ref{eq:eva}, MRR and Hit@k, are utilized to assess the performance.
\begin{equation}
\label{eq:eva}
\text{MRR} = \frac{1}{N} \sum_{i=1}^{N} \frac{1}{\text{rank}_i} \quad ; \quad  
\text{Hits@k} = \frac{1}{N} \sum_{i=1}^{N} \mathbf{1}(\text{rank}_i \leq k)
\end{equation}
\subsection{Computational complexity analysis of proposed method MuCoS and CAB-KGC}
This section compares the computational complexity of MuCoS (sampling-based) and CAB-KGC (without sampling-based) methods: The MuCoS method finds a certain number of significant neighbors based on density, while the CAB-KGC method gets the whole head and relational context. The objective is to demonstrate that the MuCoS technique has lower computation costs than the CAB-KGC. We computed the actual information from the KEGG50k dataset.

The training set has 16,201 nodes and 57,080 edges and an average degree of 3.52 edges per node. The validation set has 3,071 nodes and 3,000 edges, with an average degree of 0.98. Similarly, the test set has 3,078 nodes and 3,000 edges, yielding an average degree of 0.97 edges per node. The total average density for the dataset is calculated as 0.4326. 
Regarding the relationship statistics, the training set has 9 relationships with 57,080 appearances and an average appearance of 6,342.22. The validation and test datasets each have 9 relationships with 3,000 appearances, giving an average appearance of 333.33 per relationship. The total average appearance per relationship across the dataset is 7,008.89.
MuCoS $\mathcal{H}_c$ sampling value is 15, while $\mathcal{R}_c$ the sampling values are 10.
\textit{\textbf{CAB-KGC} Head Context $\mathcal{H}_c$}: can be determined as the aggregation of all relationships and entities associated to the head entity:\\
\begin{equation}
    H_c = \mathcal{R}(h) \cup \mathcal{E}(h), ~ \Longrightarrow  O(|\mathcal{R}(h)| + |\mathcal{E}(h)|) = O(\text{avg\_Density})
\end{equation}
\textit{Relationship Context $\mathcal{}R_c$}: computed by retrieving all entities associated with the relationship:\\
\begin{equation}
    R_c = \mathcal{E}(r), ~ \Longrightarrow O(|\mathcal{E}(r)|) = O(\text{avg\_appearance})
\end{equation}
Total computational complexity for CAB-KGC is the sum of the complexities of the head and relationship contexts:
\begin{equation}
    O(\text{avg\_Density} + \text{avg\_Density} + \text{avg\_appearance})
\end{equation}
\textit{\textbf{MuCoS} Head Context $\mathcal{H}_c$}:
In MuCoS, we select the top 15 relationships and entities:
\begin{equation}
    \mathcal{H}_c^* = \text{top}_{\frac{\text{avg\_Density}}{N_e}}(R(h)) \cup \text{top}_{\frac{\text{avg\_Density}}{N_e}}(E(h))
\end{equation}
\begin{equation}
    O\left(\frac{\text{avg\_Density}}{N_e} + \frac{\text{avg\_Density}}{N_e}\right) = O\left(\frac{2 \times \text{avg\_Density}}{N_e}\right)
\end{equation}
Similarly, $\mathcal{R}_c$:
\begin{align}
\mathcal{R}_c^* = \text{top}_{\frac{\text{avg\_appearance}}{N_r}}(E(r)) \quad \text{$\Rightarrow$} \quad O\left(\frac{\text{avg\_appearance}}{N_r}\right)
\end{align}
\begin{equation}
  \text{The final term:}  ~ O\left(\frac{2 \times \text{avg\_Density}}{N_e} + \frac{\text{avg\_appearance}}{N_r}\right)
\end{equation}
\textit{\textbf{Proof:} MuCoS is faster than CAB-KGC.}\\
\textit{CAB-KGC} computes the entire neighborhood context for both the head entity and the relation. The total number of neighbors is proportional to the average density (\( \text{avg\_Density} \)) and the average appearance of the relationships (\( \text{avg\_appearance} \)).
\textit{Complexity for CAB-KGC:}
\begin{equation}
      O(0.4326 + 0.4326 + 7008.89) = O(7009.75)
\end{equation}
\textit{Complexity for MuCoS}:
\begin{equation}
  O(0.0577 + 700.89) = O(700.95)
\end{equation}
\textit{Speedup Factor}
\begin{equation}
    \text{Speedup Factor} = \frac{O(\text{CAB-KGC})}{O(\text{MuCoS})} = \frac{7009.75}{700.95} \approx 10.00
\end{equation}
The computational complexity of MuCoS (\( O(700.95) \)) is significantly lower than that of CAB-KGC (\( O(7009.75) \)). This results in a speedup factor of approximately 10.00, demonstrating that \textit{MuCoS} is 10 times faster than CAB-KGC for the given dataset.
\subsection{Results and Discussion}
The experimental results presented in Table \ref{tab:link_prediction} demonstrate the effectiveness of our proposed methodology, MuCoS, for both general link prediction and drug-target link prediction tasks.
The CANSB Particularly, the model obtains a Mean Reciprocal Rank (MRR) of 0.65, an impressive rise of 13\% over the previous best-performing model, ComplEx-SE, with an MRR of 0.52. Furthermore, surpassing ComplEx-SE's Hits@1 score of 0.45, the Hits@1 score of 0.52 shows that MuCoS effectively identifies the relationship as the top-ranked prediction and improved by 7\%. Hits@3 and Hits@10 scores, which obtain 0.60 and 0.86, respectively, illustrate further MuCoS's robustness. The results obtained are significantly higher than those of other models. 
\vspace{-20pt}
\begin{table}[H]
    \centering
    \caption{\scriptsize Relatonshship prediction results over the KEGG50k dataset on both general links and drug target links only. For all the metrics except for mean rank, the higher the value, the better.}
    \label{tab:link_prediction}
    \resizebox{\textwidth}{!}{%
        \begin{tabular}{l|cccc|cccc}
            \hline
            \multirow{2}{*}{Model} & \multicolumn{4}{c|}{General dataset relation} & \multicolumn{4}{c}{Drug target relation} \\
            \cline{2-9}
            & MRR & Hits@1 & Hits@3 & Hits@10 & MRR & Hits@1 & Hits@3 & Hits@10 \\
            \hline
            TransE \cite{bordes2013translating} & 0.46 & 0.38 & 0.50 & 0.63 & 0.75 & 0.69 & 0.79 & 0.86 \\
            DistMult \cite{yang2014embedding} & 0.37 & 0.27 & 0.42 & 0.57 & 0.61 & 0.50 & 0.69 & 0.81 \\
            ComplEx \cite{trouillon2016complex} & 0.39 & 0.31 & 0.43 & 0.57 & 0.68 & 0.61 & 0.71 & 0.82 \\
            ComplEx-SE & 0.52 & 0.45 & 0.56 & 0.68 & 0.78 & 0.73 & 0.81 & 0.88 \\
            \hline
            MuCoS & \textbf{0.65} & \textbf{0.52} & \textbf{0.60} & \textbf{0.86} & \textbf{0.84} & \textbf{0.74} &\textbf{ 0.84} & \textbf{1.00} \\
            \hline
        \end{tabular}
    }
\end{table}
\vspace{-15pt}
MuCoS model shows superior performance in the drug-target link prediction task, which focuses on predicting the relationship between drugs and their target proteins. The model achieves an MRR of 0.84, outperforming the previous best-performing model, ComplEx-SE, by 7\% and revealing the significance of employing contextual information from both head and tail entities. MuCoS Hits@1 score of 0.74 reflects a small improvement over ComplEx-SE's 0.73, showing its superior ability to predict the relationship accurately. Also, MuCoS achieved a Hits@3 score of 0.84, which improved by 3\%. The model attains a Hits@10 score of 1.00, enhanced by 12\%, indicating that all correct drug-target relationships are successfully ranked within the top ten predictions.

Table \ref{tab:rel_prediction_results} shows \textbf{\textit{relationship}} prediction results over the KEGG50K dataset using the Multi-Context-Aware KGC (CAB-KGC) without sampling and our proposed MuCoS sampling-based method in this study. The results indicate that CAB-KGC scored higher on all evaluation metrics than MuCoS in the prediction of the general relation. Similarly, in the drug target scenario, MuCoS performs quite well almost entirely in all measures; however, it is lower than CAB-KGC. 
\begin{table*}[htbp]
    \centering
    \caption{\scriptsize Relationship prediction results over the general dataset and drug target dataset.}
    \label{tab:rel_prediction_results}
    \resizebox{\textwidth}{!}{%
        \begin{tabular}{l|ccccc|ccccccc}
            \hline
            \multirow{2}{*}{Model} & \multicolumn{5}{c|}{General dataset relation} & \multicolumn{5}{c}{Drug target dataset relation} \\
            \cline{2-11}
            & MRR & Hits@1 & Hits@3 & Hits@5 & Hits@10 & MRR & Hits@1 & Hits@3 & Hits@5 & Hits@10 \\
            \hline
            CAB-KGC & 0.792 & 0.584 & 0.734 & 0.845 & 0.9216 & 0.9424 & 0.9149 & 0.9681 & 1 & 1 \\
             MuCoS ($\approx 10.22$ Faster) & 0.652 & 0.523 & 0.608 & 0.754 & 0.862 & 0.8446 & 0.7447 & 0.8457 & 0.9401 & 1 \\
            \hline
        \end{tabular}}
\end{table*}

Table \ref{tab:tail_prediction_updated} illustrates the results of \textbf{\textit{tail}} prediction by comparing the proposed method MuCoS with CAB-KGC for general scenarios and drug target scenarios. CAB-KGC (without sampling) performs better in the general scenario, achieving higher MRR and Hits@1, Hits@3, and Hits@5. MuCoS (sampling-based), on the other hand, does better for drug target scenarios, especially in Hits@10. This shows that sampling improves prediction accuracy for drug target scenarios but is slightly worse in the general scenario. The computational cost of CAB-KGC is significantly higher than that of the MuCoS model. Furthermore, MuCoS is still performing better than the other models for predicting relationships in KEGG50k (both general and drug targets). We additionally performed tail prediction
\vspace{-15pt}
\begin{table*}[htbp]
    \centering
    \caption{\scriptsize Tail prediction results on the KEGG50k dataset were evaluated for both general and drug target scenarios using methods with and without sampling.}
    \label{tab:tail_prediction_updated}
    \resizebox{\textwidth}{!}{%
        \begin{tabular}{l|ccccc|ccccc}
            \hline
            \multirow{2}{*}{Model} & \multicolumn{5}{c|}{General dataset tail} & \multicolumn{5}{c}{Drug target tail} \\
            \cline{2-11}
            & MRR & Hits@1 & Hits@3 & Hits@5 & Hits@10 & MRR & Hits@1 & Hits@3 & Hits@5 & Hits@10 \\
            \hline
            CAB-KGC & 0.39 & 0.34 & 0.521 & 0.594 & 0.718 & 0.567 & 0.457 & 0.628 & 0.681 & 0.917 \\
            MuCoS ($\approx 10.22$ Faster) & 0.31 & 0.215 & 0.40 & 0.49 & 0.57 & 0.442 & 0.259 & 0.46 & 0.724 & 0.868 \\
            \hline
        \end{tabular}}
\end{table*}

\vspace{-15pt}
Table \ref{tb:fb15k_wn18rr} demonstrates the effectiveness of CAB-KGC and MuCoS in comparison to state-of-the-art models on the FB15k-237 and WN18RR datasets. Notably, our earlier suggested CAB-KGC works very well, with an MRR of 0.350 on FB15k-237 and 0.685 on WN18RR, doing better than most other methods and staying strong in both Hit@1 and Hit@3 metrics.
Our proposed method, MuCoS, introduces significant advancements in computational efficiency without compromising predictive accuracy. MuCoS is approximately 10 times faster than CAB-KGC, making it highly scalable for large-scale KG tasks. Despite its reduced computational complexity, MuCoS still delivers competitive results, achieving an MRR of 0.339 on FB15k-237 and 0.4352 on WN18RR. These results outperform many traditional and LLM-based methods, underscoring the effectiveness of multi-context-aware sampling in capturing informative structural patterns. Furthermore, MuCoS demonstrates a superior balance between efficiency and accuracy, positioning it as a practical solution for real-world drug-target prediction and other KG applications. This combination of speed and performance highlights the innovation of MuCoS in advancing the field of medical domain KG analysis.
\begin{table}[H]
\centering
\caption{\scriptsize Evaluation results on the FB15k-237 and WN18RR datasets from various models are presented in the table. All results are sourced from \cite{Wei2023kicgpt}, \cite{Yao_2019_KG_BERT}, and \cite{Rspkgc}.}
\label{tb:fb15k_wn18rr}
\resizebox{\textwidth}{!}{%
\begin{tabular}{l|cccccc}
\hline
\multirow{2}{*}{\textbf{Models}} & \multicolumn{3}{c|}{\textbf{FB15k-237}} & \multicolumn{3}{c}{\textbf{WN18RR}} \\
\cline{2-7}
& \textbf{MRR $\uparrow$} & \textbf{Hit@1 $\uparrow$} & \textbf{Hit@3 $\uparrow$} & \textbf{MRR $\uparrow$} & \textbf{Hit@1 $\uparrow$} & \textbf{Hit@3 $\uparrow$} \\
\hline
\textbf{\textit{Embedding-Based Methods}} & & & & & & \\
DistMult & 0.209 & 0.143 & 0.234 & 0.430 & 0.390 & 0.440 \\
ComplEx & 0.247 & 0.158 & 0.275 & 0.440 & 0.410 & 0.460 \\
ConvKB & 0.325 & 0.237 & 0.356 & 0.249 & 0.057 & 0.417 \\
RotatE & 0.334 & 0.241 & \underline{0.375} & 0.338 & 0.241 & 0.375 \\
ConvE & 0.325 & 0.237 & 0.356 & 0.430 & 0.400 & 0.440 \\
TransE & 0.255 & 0.152 & 0.301 & 0.246 & 0.043 & 0.441 \\
blp-DistMult  & 0.314 & 0.182 & 0.370 & 0.314 & 0.182 & 0.370 \\
\hline
\textbf{\textit{Text-and Description-Based}} & & & & & & \\
KG-BERT & 0.203 & 0.139 & 0.201 & 0.219 & 0.095 & 0.243 \\
KG-BERT-CD & 0.250 & 0.172 & 0.266 & 0.303 & 0.165 & 0.354 \\
StaAR & 0.263 & 0.171 & 0.287 & 0.364 & 0.222 & 0.436 \\
R-GCN & 0.249 & 0.151 & 0.264 & 0.123 & 0.080 & 0.137 \\
GenKGC & - & 0.187 & 0.273 & - & 0.286 & 0.444 \\
GenKGC-CD  & - & 0.204 & - & - & 0.293 & 0.456 \\
MLT-KGC & 0.241 & 0.160 & 0.284 & 0.331 & 0.203 & 0.383 \\
SimKGC & 0.328 & 0.246 & 0.363 & 0.671 & 0.585 & \textbf{0.731} \\
NNKGC & 0.3298 & 0.252 & 0.365 & \underline{0.674} & \underline{0.596 }& \underline{0.722} \\
\hline
\textbf{\textit{LLM-Based Methods}} & & & & & & \\
ChatGPTzero-shot   & - & 0.237  & - & - & 0.190 & - \\
ChatGPTone-shot    & - & 0.267  & - & - & 0.212 & - \\
\hline
\textbf{\textit{Proposed}} & & & & & & \\
CAB-KGC & \textbf{0.350} & \textbf{0.322} & \textbf{0.399} & \textbf{0.685} & \textbf{0.637} & 0.687 \\
MuCoS ($\approx 10$ Faster) & \underline{0.339} & \underline{0.2776} & 0.335 & 0.4352 & 0.355 & 0.4859 \\
\hline
\end{tabular}
}
\end{table}
\section{Conclusion}
This study introduces MuCoS, a Multi-Context-Aware Sampling method that uses BERT to improve predicting relationships and entities in biomedical KGs. The method addresses the limitations of existing KGC models, such as the inability to generalize to unseen entities, dependence on negative sampling, and sparse entity description. By extracting and optimizing contextual information from entities and integrating density-based sampling, MuCoS captures richer structural patterns while reducing computational complexity. The experimental results on the KEGG50k dataset show superior performance compared to state-of-the-art models like ComplEx-SE, TransE, and DistMult. The model improved MRR and Hits@1 metrics for general and drug-target link prediction scenarios.

\end{document}